\documentclass{article}

\PassOptionsToPackage{numbers, compress}{natbib}

\usepackage[final]{nips_2018}

\usepackage[utf8]{inputenc} 
\usepackage[T1]{fontenc}    
\usepackage{hyperref}       
\usepackage{url}            
\usepackage{booktabs}       
\usepackage{amsfonts}       
\usepackage{nicefrac}       
\usepackage{microtype}      

\usepackage[dvipsnames]{xcolor}
\usepackage{color}

\definecolor{redcolor}{RGB}{219, 0, 0}
\definecolor{bluecolor}{RGB}{0, 0, 219}

\usepackage{placeins}

\usepackage{tikz}
\usetikzlibrary{positioning}
\usepackage{algorithm}
\usepackage{algorithmic}

\usepackage[subtle,bibbreaks=normal,mathspacing=normal]{savetrees}

\usepackage{subcaption} 
\usepackage[font=small]{caption}
\usepackage{amsmath}
\usepackage{amsfonts}

\usepackage[colorinlistoftodos,prependcaption,textsize=tiny]{todonotes}

\usepackage{resizegather}
\usetikzlibrary{bayesnet}
\usepackage{rotating}

\usepackage{easytable}
\usepackage[toc,page]{appendix}
\usepackage{float}
\usepackage{resizegather}
\usepackage{array}
\usepackage{balance}
\usepackage{flushend}

\graphicspath{{images/}}

\usepackage{xspace}
\makeatletter
\DeclareRobustCommand\onedot{\futurelet\@let@token\@onedot}
\def\@onedot{\ifx\@let@token.\else.\null\fi\xspace}
\def\eg{\emph{e.g}\onedot} 
\def\ie{\emph{i.e}\onedot}

\def\wrt{w.r.t\onedot} 

\makeatother

\newcommand{\y}{\mathbf{y}}
\newcommand{\x}{\mathbf{x}}
\newcommand{\s}{\mathbf{s}}
\newcommand{\uu}{\mathbf{u}}

\newcommand{\DP}[1]{\mathcal{DP}\!\left(#1\right)}

\usepackage{titlesec}
\titlespacing*{\paragraph}{0pt}{0.0ex plus .2ex}{1em}

\title{Sequence Alignment with Dirichlet Process Mixtures}

\author{%
\begin{tabular}{cc}
\begin{tabular}[t]{c}
\vspace{0.2cm} \\
Ieva Kazlauskaite \\
\normalfont Department of Computer Science \\ 
\normalfont University of Bath \\
\normalfont UK \\
\texttt{i.kazlauskaite@bath.ac.uk}
\end{tabular} &
\begin{tabular}[t]{c}
\vspace{0.2cm} \\
Ivan Ustyuzhaninov \\
\normalfont Centre for Integrative Neuroscience\\
\normalfont University of T\"ubingen\\
\normalfont Germany
\end{tabular} \\
\addlinespace[4ex] 
\begin{tabular}[t]{c}
Carl Henrik Ek \\
\normalfont Faculty of Engineering \\
\normalfont University of Bristol \\
\normalfont UK
\end{tabular} &
\begin{tabular}[t]{c}
Neill D. F. Campbell \\
\normalfont Department of Computer Science \\
\normalfont University of Bath \\
\normalfont UK
\end{tabular}
\vspace{-0.5cm}
\end{tabular}
}

\begin{document}

\maketitle

\begin{abstract}
  We present a probabilistic model for unsupervised alignment of high-dimensional time-warped sequences based on the Dirichlet Process Mixture Model (DPMM). We follow the approach introduced in \cite{Kazlauskaite:2018} of simultaneously representing each data sequence as a composition of a true underlying function and a time-warping, both of which are modelled using Gaussian processes (GPs)~\cite{Rasmussen:2005}, and aligning the underlying functions using an unsupervised alignment method. In \cite{Kazlauskaite:2018} the alignment is performed using the GP latent variable model (GP-LVM)~\cite{Lawrence:2005vk} as a model of sequences, while our main contribution is extending this approach to using DPMM, which allows us to align the sequences temporally and cluster them at the same time. We show that the DPMM achieves competitive results in comparison to the GP-LVM on synthetic and real-world data sets, and discuss the different properties of the estimated underlying functions and the time-warps favoured by these models.
\end{abstract}

\section{Introduction}
\label{intro}
For many machine learning tasks we are faced with the scenario of learning a model from non-stationary sequential data that results from studying several views of the same underlying phenomenon.
Consider the following examples: replicated scientific experiments often vary in timing across different trials, different subjects might require different amounts of time to perform the same task, the peaks of the neuronal action potential waveforms do not match each other perfectly, telecommunications suffer from temporal jitter, climate patterns are often cyclic though particular events take place at slightly different times in the year.
However, most sample statistics, such as mean and variance, are designed to capture variations in amplitude rather than phase or timing; this results in increased sample variance, blurred fundamental data structures and an exaggerated number of principal components necessary to describe the data.
Therefore, the alignment of data is often performed as a non-trivial pre-processing step.

Traditionally, the notion of sequence correspondence is defined using a measure of pairwise similarity integrated across the sequences~\cite{Berndt:1994, Keogh:2001, Drydmard:2016, Hsu:2005, Zhou:2009, Zhou:2012}. 
We build on~\cite{Kazlauskaite:2018} where the models of the individual sequences and the alignment across sequences are cast within a single framework. 
Differently from the commonly used approach of interpolating between observations, the underlying functions and the warps are modelled using GPs which jointly regularise the solution space of the ill-posed alignment problem. The alignment is performed using dimensionality reduction which preserves similarities in the observation space while imposing the preference for dissimilar data points to be placed far apart in the latent space. In particular, the authors propose to use a GP-LVM that places independent GPs over the data features and optimises the corresponding latent variables.

Similarly to~\cite{Kazlauskaite:2018}, we use GPs to model the data and the warps, which allows us to reject the observation noise in a principled manner, and imposes a smoothness constraint on the warping functions. In contrast, however, we consider a Dirichlet process mixture model~\cite{Ferguson:1973} as a model for alignment. The DPMM performs clustering explicitly and allows us to automatically find the optimal number of underlying functions explaining the observed sequences.


\section{Methodology} \label{sec:methodology}

Let us assume that we are given $J$ noisy observed sequences $\{\y_j\}_{j=1}^J$ where each sequence comprises of $N$ time samples $\y_j := (y_{j,1}, \ldots, y_{j,N}) \in \mathbb{R}^N$. We consider each sequence to be generated by sampling a latent function $f_j(x)$ at points $\{g_j(x_n)\}_{n=1}^N$, \ie $y_{j,n} = f_j( \, g_j(x_n) ) + \varepsilon_j$, where $\x := (x_1, \ldots, x_N) \in \mathbb{R}^N$ are known evenly-spaced inputs (same for all sequences), $\{g_j(\cdot)\}_{j=1}^J$ are the warping functions (different for different sequences) and $\varepsilon_j \sim \mathcal{N}(0, \beta_j^{-1})$ is i.i.d. Gaussian noise.

We are interested in the case where the number of distinct latent functions $f_j(\cdot)$ is smaller than $J$, \ie some of the observed sequences are generated from the same underlying functions. In this setting, the observed sequences were \emph{misaligned} by applying different time warpings $g_j(\cdot)$ to the input locations. We treat the \emph{aligned} sequences $\: \s_j := (f_j(x_1), \ldots, f_j(x_N)) \in \mathbb{R}^N$ as if they were observed, and hence we refer to $\{\s_j\}$ as pseudo-observations.
The rest of this section is structured as follows: first, we briefly describe the model over individual sequences $\{\y_j\}$, which is the same as in \cite{Kazlauskaite:2018}, and we refer the reader to this work for more details; next, we introduce an aligning objective based on the DPMM.

\paragraph{Model over time} We model latent functions $f_j(\cdot)$ as stationary GPs, $f_j \sim \mathcal{GP}(0, k_{\theta_n}\!(\cdot, \cdot))$ with a squared exponential kernel. Denoting the finite-dimensional evaluation of the warping function as $G_j := (g_j(x_1), \ldots, g_j(x_N))$, the joint likelihood of observed and aligned sequences is
\begin{equation}
p\!\left(\begin{bmatrix} \s_j \\[5pt] \y_j\end{bmatrix}  \middle| \, G_j, \x, \theta_j \right) \sim 
\mathcal{N}\left(0, 
    \begin{bmatrix}
        k_{\theta_j}\!(\x, \x) & k_{\theta_j}\!(\x, G_j) \\[5pt]
         k_{\theta_j}\!(G_j, \x) & k_{\theta_j}\!(G_j, G_j)
    \end{bmatrix} + \beta_j^{-1}I
\right)\ .
\label{eq:gp-likelihood}
\end{equation}

We encode our preference for smooth warping functions by also modelling $g_j(\cdot)$ as a GP with a smooth kernel function.  
The $\{G_j\}$ are constrained to be monotonic in the range $[-1,1]$ using a cumulative sum of the set of auxiliary variables $\uu_j \in \mathbb{R}^{N}$ with a softmax renormalisation.
The distribution over $G_j$ is $p(G_j \mid \x, \omega) \sim \mathcal{N}(0, k_{\omega_j}\!(X, X))$ with hyperparameters $\omega_j$.

\paragraph{Alignment model} So far we have modelled each sequence $\y_j$ in isolation, and there is nothing in the model that encourages it to use the smallest possible number of distinct latent functions $\{f_j(\cdot)\}$ to explain the data. One way to introduce such a constraint is to add a regularisation term that encourages clustering of aligned sequences $\{\s_j\}$ into a small number of clusters. Indeed, the subset $\{\s_{c_j}\}$ of the aligned sequences that belong to the same cluster would be similar to each other, meaning that the corresponding latent functions $\{f_{c_j}\!(\cdot)\}$ are also similar when evaluated at $\x$, while the GP prior on $\{f_{c_j}\!(\cdot)\}$ enforces that the functions are smooth and nearby values are correlated.

We propose clustering $\{\s_j\}$ using the DPMM. As a non-parametric mixture model, DPMM allows us to automatically infer the number of clusters (\ie distinct aligned sequences) from the data.
The locations of the mixture components and the mixing weights are distributed according to a sample from a Dirichlet process, and, depending on the parameters of the process, such a prior seeks to explain the data using only a few mixture components, corresponding to aligning the underlying latent functions.
Formally, the DPMM can be defined as follows:
\begin{equation}
    G \sim \DP{\alpha, G_0}, \quad \eta_j \sim G, \quad  \s_j \sim p(\s_j | \eta_j),
\end{equation}
where $G$, a sample from the Dirichlet process, can be thought of as an infinite weighted sum of delta functions at locations sampled from the base distribution $G_0$ (\eg a Gaussian). The scaling parameter $\alpha$ controls the entropy of those weights; for small values of $\alpha$ only a few weights are significantly above 0, meaning that the data $\s_j$ are sampled from one of a few mixture distributions with parameters $\eta_j$. An explicit construction of the DPMM is based on a stick breaking representation:
\begin{equation}
\begin{aligned}
    v_i & \sim \text{Beta}(1, \alpha), \, i = \{1,2,3,\ldots\}, \\
    \pi_i & = v_i (1 - v_1)  \ldots (1 - v_{i-1}), \\
    \eta_i & \sim G_0, \quad
    z_i \sim \text{Mult}(\pi), \quad
    \s_j \sim p(\s_j | \eta_{z_j}).
\end{aligned}    
\end{equation}
We use DPMM to regularise the GPs that are fitted to the data, and hence we optimise the data likelihood of the DPMM jointly with the GP models. Since the data likelihood is not available in closed form, we approximate it with a variational lower bound \cite{Blei:2005}. Specifically, we approximate posterior distributions over $v_i$, $\eta_i$ and $z_j$ with factorised Beta, Gaussian and Multinomial distributions respectively:
\begin{equation}
    q(\boldsymbol{v}, \boldsymbol{\eta}, \boldsymbol{z}) = 
    \prod\limits_{t=1}^{T-1}q_{\gamma_t}(v_t)
    \prod\limits_{t=1}^{T}q_{\tau_t}(\eta_t)
    \prod\limits_{j=1}^{J}q_{\phi_j}(z_j),
    \label{eq:dpmm-likelihood}
\end{equation}
where $T$ is the maximal number of clusters in the mixture (the infinite mixture model is truncated for the variational approximation). This approximation allows us to obtain a lower bound on the data likelihood:
\begin{equation}
    \log p(\boldsymbol{s}_j) \geq \mathcal{L}_q(\boldsymbol{s}_j, \boldsymbol{v}, \boldsymbol{\eta}, \boldsymbol{z}) := \mathbb{E}_q[ \log p(\boldsymbol{s}_j, \boldsymbol{v}, \boldsymbol{\eta}, \boldsymbol{z})] - \mathbb{E}_q[ \log q(\boldsymbol{v}, \boldsymbol{\eta}, \boldsymbol{z})],
\end{equation}
where both terms are analytically tractable, and we refer the reader to \cite{Blei:2005} for their exact expressions.


\paragraph{Learning}

We want the observed data $\{\y_j\}$ and aligned sequences $\{\s_j\}$ to be modelled by the GPs (by $f(g(\x))$ and by $f(\x)$ respectively), and the aligned sequences $\{\s_j\}$ to be clustered into groups, therefore, we simultaneously maximise the GP data likelihood \eqref{eq:gp-likelihood} and the lower bound on the DPMM likelihood \eqref{eq:dpmm-likelihood}. The GP likelihood \eqref{eq:gp-likelihood} includes the evalution of the warping GP, $G_j$, which we cannot integrate out analytically, therefore, we use a point estimate by including the likelihood of $G_j$ in the objective, and directly optimising $\{\uu_j\}$, which parametrise $G_j$. Overall, the optimisation objective is
\begin{equation}
    \mathcal{J}(\s, G, \boldsymbol{v}, \boldsymbol{\eta}, \boldsymbol{z}) = \sum\limits_{j=1}^J \left[
    p\!\left(\begin{bmatrix} \s_j \\[5pt] \y_j\end{bmatrix}  \middle| \, G_j, \x, \theta_j \right) + 
    \mathcal{L}_q(\boldsymbol{\s}_j, \boldsymbol{v}, \boldsymbol{\eta}, \boldsymbol{z}) + p(G_j \mid \x, \omega)
    \right],
\end{equation}
which we maximise \wrt the pseudo-observations of the aligned sequences $\{\s_j\}$, the DPMM variational parameters $\boldsymbol{v}, \boldsymbol{\eta}, \boldsymbol{z}$, the auxilary variables of the warping functions $\{\uu_j\}$, and the hyper-parameters of the GPs and the DPMM. Among the hyper-parameters of the DPMM are the scaling parameter $\alpha$, variance of the base distribution $G_0$ (we assume it is a zero-mean Gaussian), and the variance of mixture components (we assume diagonal-covariance Gaussians with the same variance for all components). The scaling parameter $\alpha$ and the variance of $G_0$ are directly optimised (we include a Gamma prior on them), while the variance of mixture components is set to $1 / \beta$, \ie the estimated noise in the GP fits, which we assume to be the same for all sequences.

\section{Experiments} \label{sec:experiments}
\paragraph{Synthetic dataset} We consider a set of 10 sequences generated using $\mathrm{sinc}(\x)$ and $\x^3$, where $\x$ is a linearly spaced vector of values in $[-1,1]$, and warped using randomly generated monotonically increasing warping functions. We define (1) the mean (median) alignment error as the sum of means (medians) of pairwise distances between observations within each group of sequences in the N-dimensional space, (2) the data fit as the standard deviation of the estimated observational noise ($\sqrt{1 / \beta}$), and (3) the warping complexity as the sum of the absolute values of differences between components of $\uu_j$, which corresponds to the total variation of variables that define the warps. We provide a quantitative comparison between our method and the method that uses GP-LVM~\cite{Kazlauskaite:2018} in terms of the three criteria in Fig.~\ref{fig:toy_err} (see Appendix) as a function of a warping parameter $\omega$ where 0 corresponds to no warping and the warps get progressively larger as $\omega$ increases. Our model achieves lower alignment error in situations where $\omega$ is small or intermediate while the GP-LVM achieves smaller mean alignment errors for very large warps. However, the median errors of our model stay low even for high values of $\omega$. An example of this behaviour is provided in Fig.~\ref{fig:compare} (see Appendix). If one of the sequences is an outlier due to a large warp, the DPMM tends to create a new cluster for it, while the GP-LVM favours the solution that recovers the two groups of sequences but is unable to align the sequences within the two groups accurately.

\begin{figure}[h!]
\centering
        \begin{subfigure}[h]{1.\textwidth}
            \centering
             \includegraphics[width=0.985\textwidth]{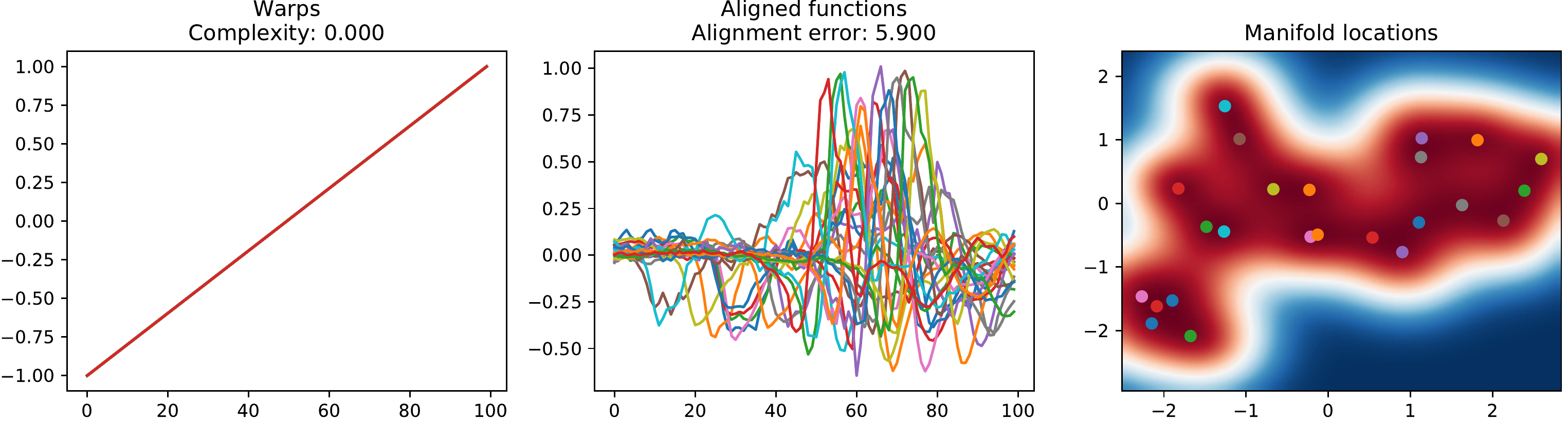} 
        \caption{Input sequences. The corresponding clustering does not discover the two different types of heartbeats.} \end{subfigure} 
        \begin{subfigure}[h]{1.\textwidth}
               \centering \includegraphics[width=0.985\textwidth]{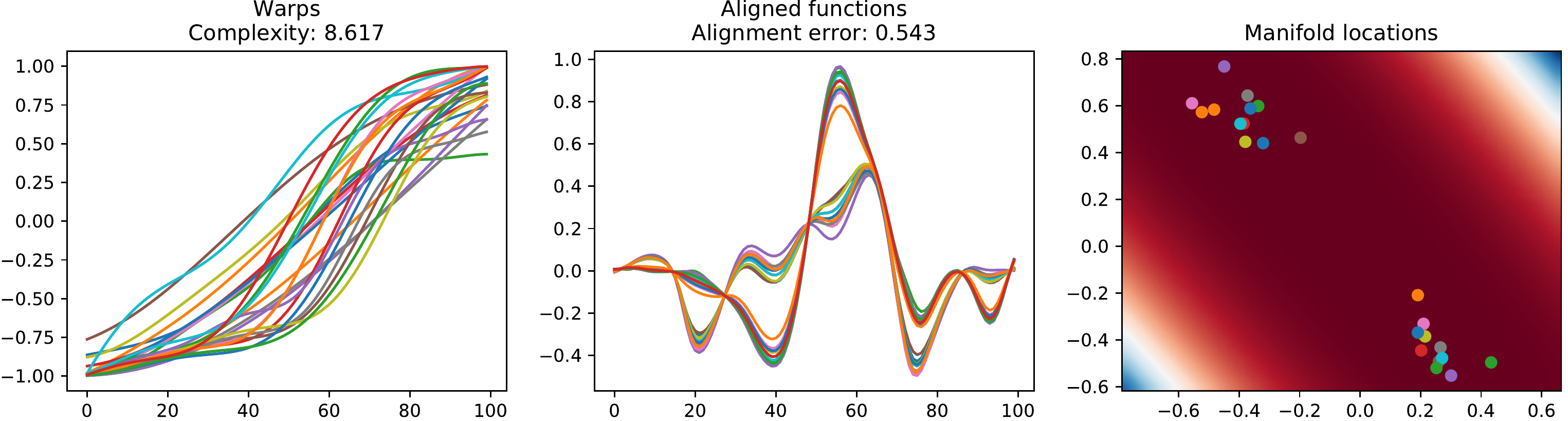} 
        \caption{GP-LVM alignment~\cite{Kazlauskaite:2018}.} \end{subfigure}
        \begin{subfigure}[h]{1.\textwidth}
               \centering \includegraphics[width=0.985\textwidth]{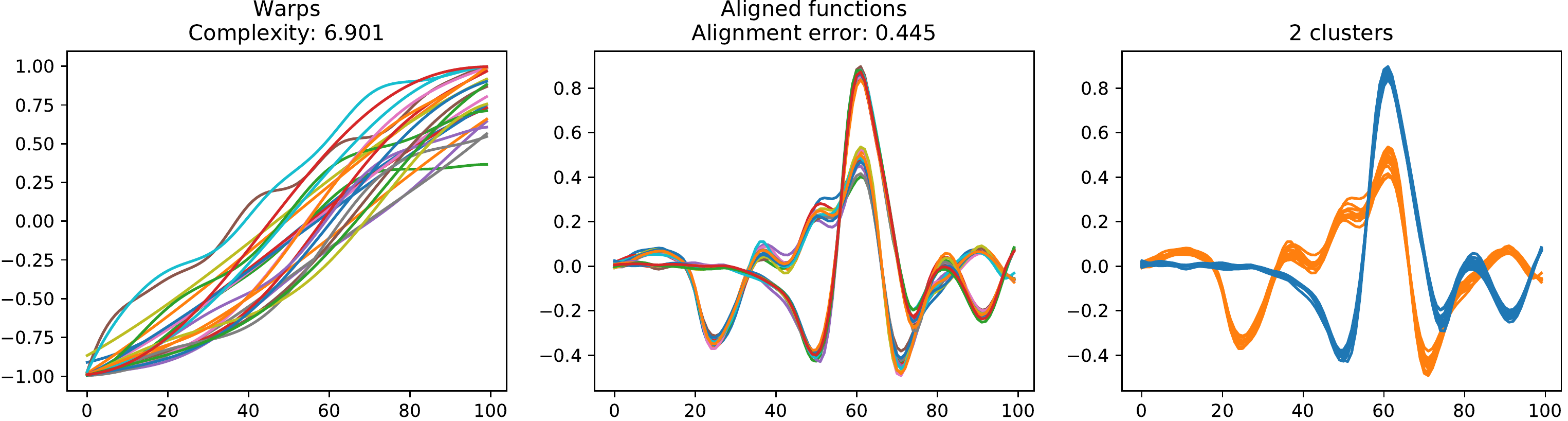} 
        \caption{DPMM alignment.} \end{subfigure}
        \caption{Alignment of heartbeats data~\cite{Bentley:2011}.}\label{fig:heartbeats}
\end{figure}

\paragraph{Heartbeats data} We consider a data set of heart beat sounds, which have a clear "lub dub, lub dub” pattern that varies temporally depending on the age, health, and state of the subject~\cite{Bentley:2011}. Similarly to~\cite{Kazlauskaite:2018}, our model is able to automatically align and cluster the heart sounds into two distinct patterns. In both approaches a Mat\'{e}rn $3/2$ kernel is used, that takes into account the rapid variations in the recordings while also limiting the effect of the uninformative high frequency noise. Fig.~\ref{fig:heartbeats} gives a comparison of the alignment and the clustering of a set of heartbeats. Our model achieves a smaller alignment error and estimates simpler warps (closer to an identity function). 

\section{Discussion} 
\label{sec:discussion}
We have presented a probabilistic model that is able to implicitly align inputs that contain temporal variations. Our approach builds on~\cite{Kazlauskaite:2018}, where we replace the latent variable model with a DPMM. Our model performs explicit clustering in the data space (and automatically estimates the number of clusters), while the GP-LVM performs dimensionality reduction and looks for a set of sequences that exhibit a simple structure in the latent low dimensional space (and allows to sample new sequences). 
While the experimental results suggest that both models perform well on real and synthetic data sets, they display different behaviour. The GP-LVM  aligns the sequences constrained to them staying on a low-dimensional structure; in comparison, the DPMM does not have this global constraint and aligns sequences in each estimated cluster independently from the other clusters. In future work we propose developing a model which makes use of the global low-dimensional structure and the unconstrained alignment within clusters. The main limitation of our method is related to the sensitivity to the choice of hyper-parameters in the variational approximation of the lower bound on the DPMM likelihood, and one possible direction to overcome this limitation could be to perform variational inference over the hyper-parameters instead of optimising them directly.
A successful application of our method on new data sets relies on the choice of priors for alignment, and we propose to explore the effect of using alternative prior processes, such as the Pitman-Yor process.
\section*{Acknowledgements}
This work has been supported by EPSRC CDE (EP/L016540/1) and CAMERA (EP/M023281/1) grants.
\bibliography{references}
\bibliographystyle{plain}

\newpage
\appendix

\renewcommand{\thesection}{Appendix:}
\renewcommand\thefigure{A\arabic{figure}}    
\setcounter{figure}{0} 

\section{Experiments on the synthetic data}
\label{sec:appendix-synthetic-data}

\begin{figure}[h!]
\centering
        \begin{subfigure}[h]{1.\textwidth}
                \includegraphics[width=\textwidth]{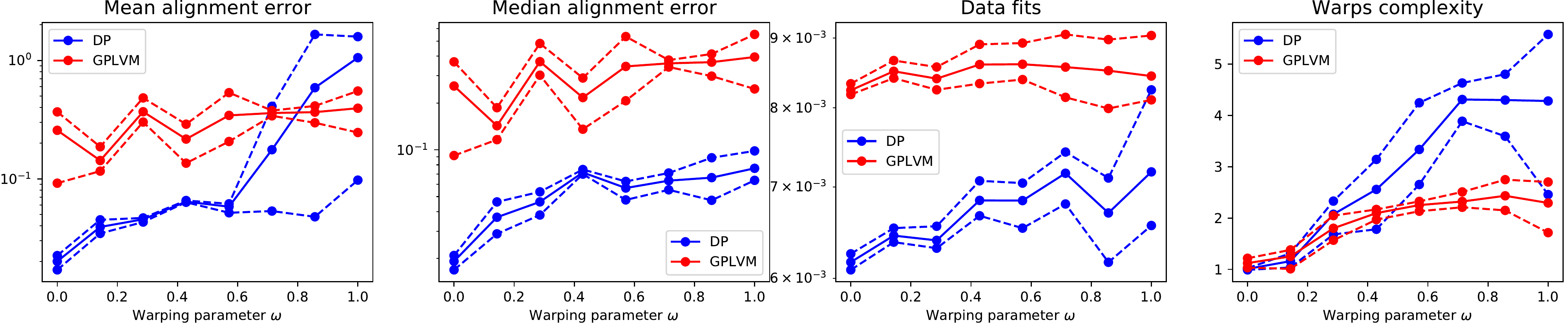}
         \end{subfigure} 
         \caption{Comparison of alignment error, data fit and warp complexity for our model and~\cite{Kazlauskaite:2018}. Solid lines show mean values across 5 trials, dashed lines show min and max values.}\label{fig:toy_err}
\vspace{0.5cm}
        \begin{subfigure}[h]{1.0\textwidth}
            \includegraphics[width=\textwidth]{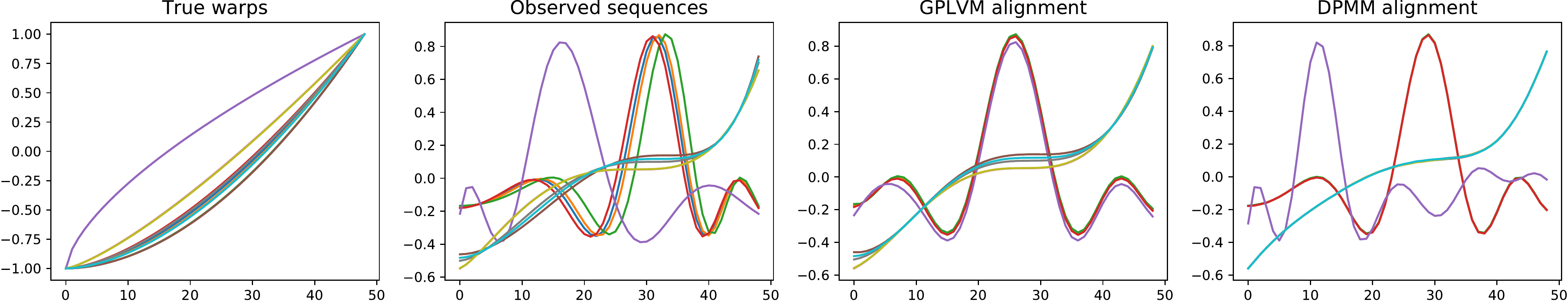}
        \end{subfigure}
        \caption{An example of behaviour for~\cite{Kazlauskaite:2018} and our model for the warping parameter $\omega$ of 1.0 in Fig.~\ref{fig:toy_err}.} \label{fig:compare}
\end{figure}

\end{document}